\documentclass{article}
\pdfoutput=1
\usepackage{ijcai24}

\usepackage{times}
\usepackage{soul}
\usepackage{url}
\usepackage[hidelinks]{hyperref}
\usepackage[utf8]{inputenc}
\usepackage[small]{caption}
\usepackage{graphicx}
\usepackage{amsmath}
\usepackage{amsthm}
\usepackage{booktabs}
\usepackage{algorithm}
\usepackage{algorithmic}
\usepackage[switch]{lineno}
\usepackage{tabularx}
\usepackage{amsfonts}

\title{Correct \emph{and} Optimal: the Regular Expression Inference Challenge}
\author{%
    Mojtaba Valizadeh$^1$\textsuperscript{\footnote{equal contribution}} \and%
    Philip John Gorinski$^{2}$\textsuperscript{$^{\ast}$} \and%
    Ignacio Iacobacci$^2$ \And%
    Martin Berger$^{1,2,3}$\textsuperscript{\footnote{contact author\\Extended version including Technical Appendix of paper published at IJCAI2024, available at \url{https://ijcai24.org/}}}
    \\
    \affiliations {%
        $^1$University of Sussex\quad
        $^2$Huawei Noah's Ark Lab, London\quad
        $^3$Montanarius Ltd
    }
    \emails{\small
        Valizadeh.Mojtaba@gmail.com, %
        p.j.gorinski@gmail.com, ignacio.iacobacci@huawei.com, %
        contact@martinfriedrichberger.net
    }
}

\newcommand{\NI}{\noindent}
\newcommand{\EG}{e.g.,\ } 
\newcommand{\IE}{\textit{i.e.},\ } 
\newcommand{\WRT}{w.r.t.\ } 
\newcommand{\EMPH}[1]{\emph{#1}}
\newcommand{\FS}{\rightarrow}
\newcommand{\NAT}{\mathbb{N}}

\newcommand{\NOVSPACEPARAGRAPH}[1]{\NI\textbf{\emph{#1}.}}
\newcommand{\PARAGRAPH}[1]{\vspace{2mm}\NOVSPACEPARAGRAPH{#1}}

\newcommand{\INTERSECT}{\mathbin{\&}}
\newcommand{\LANG}[1]{\mathsf{Lang}(#1)}

\newcommand{\RE}[1]{\mathsf{RE}(#1)}
\newcommand{\COST}[1]{\mathsf{cost}(#1)}

\newcolumntype{Y}{>{\centering\arraybackslash}X}

\begin{document}

\maketitle

\begin{abstract}
We propose \emph{regular expression inference (REI)} as a challenge for code/language modelling, and the wider machine learning community. REI is a supervised machine learning (ML) and program \emph{optimisation} task, and poses the problem of finding minimal regular expressions from examples: Given two finite sets of strings $P$ and $N$ and a cost function $\COST{\cdot}$, the task is to generate an expression $r$ that accepts all strings in $P$ and rejects all strings in $N$, while no other such expression $r'$ exists with $\COST{r'}<\COST{r}$.
REI has advantages as a challenge problem: (i) regular expressions are well-known, widely used, and a natural idealisation of code; (ii) REI's asymptotic worst-case complexity is well understood; (iii) REI has a small number of easy to understand parameters (e.g.~$P$ or $N$ cardinality, string lengths of examples, or the cost function); this lets us easily finetune REI-hardness; (iv) REI, with its emphasis on optimisation, is an unsolved problem for deep learning based ML.
Recently, an REI solver was implemented on GPUs, using program synthesis techniques. This enabled, for the first time, fast generation of minimal regular expressions for complex REI instances. Building on this advance, we generate and publish
the first large-scale datasets for REI, and devise and evaluate several initial heuristic and machine learning baselines.
We invite the community to participate and explore ML methods that learn to solve REI problems. We believe that progress in REI directly translates to progress in code/language modelling.
\end{abstract}

\section{Introduction}\label{introduction}
We propose regular expression inference (REI) as a challenge for machine learning (ML) communities.

Regular expression inference is the task of finding a regular expression (RE) $r$ given a positive set of strings \EMPH{P} and negative set of strings \EMPH{N} as well as a cost function $\COST{\cdot}$, such that $r$ is \EMPH{precise} -- it accepts all strings in P while rejecting all strings in N -- and \emph{minimal} \WRT the cost function.

We assume the reader is familiar with REs; here is an example of one that specifies the language of all strings with characters from the
alphabet $\{0, 1\}$ that start with 10:
{\small{\[
   10(0 + 1)^*
\]}}

The exact nature of $\COST{\cdot}$ will play a key role later, for now a naive
understanding of cost, for example the length of $r$ as a string, is
sufficient. 

Regular expression inference always has a trivial solution. Consider inferring a
regular expression from
{\small{
\begin{align*}
\text{P:} &\quad 10, \ 101, \ 100, \ 1010, \ 1011, \ 1000, \ 1001 \\
\text{N:} & \quad \epsilon, \ 0, \ 1, \ 00, \ 11, \ 010
\end{align*}
}}
Clearly, the union of all the positive examples
{\small{
\begin{align*}
   10 + 101 + 100 + 1010 + 1011 + 1000 + 1001
\end{align*}
}}
is correct but trivial, and can be seen as \EMPH{overfitting}. 
The problem becomes highly non-trivial
if we ask for a \EMPH{minimal} RE such as
$
10(0 + 1)^*
$,
assuming a uniform cost function.

In this work, we present the regular expression inference challenge (REIC) as an open-ended challenge to address REI with machine learning/deep learning. We detail the precise nature of REIC, introduce large-scale datasets suitable for supervised training and evaluation, and present baselines including the use of an instruction-tuned pretrained very large code language model, and a first supervised approach.
To our knowledge, we are the first to put REI forward as an
organised, well-defined and automatically scored challenge.
All data and starter code to recreate our baselines is provided via CodaLab \cite{codalabcompetitions} on
the REIC site:
\begin{center}
\url{https://codalab.lisn.upsaclay.fr/competitions/15096}
\end{center}

\section{Background \& Related Work}
Regular Expressions are well-known mathematical
structures, invented in the context of modelling biological neurons
\cite{KleeneSC:repeveinnafa,HopcroftJE:intauttlac}.  Abstractly, REs are a constrained
mechanism for succinct, finite specifications of finite and infinite
languages. While all finite languages are definable by REs, they can only specify simple infinite ones.  

REs are equivalent not only to regular languages, the least
expressive language type in the Chomsky-Sch\"utzenberger hierarchy
\cite{chomsky1956three}, but also finite state automata, whether deterministic
(DFAs) or non-deterministic (NFAs), as well as many other important
formalisms such as read-only Turing machine \cite{turing1936computable} and monadic second-order logic \cite{10.5555/2414243}.

REI is a special case of grammar inference (GI), a long-standing research subject in AI.
GI emerged from Chomsky-style linguistics, which models the problem of
human language learning: how can children infer the grammar of
their native language from a small number of examples?  A simple model of language learning from examples is the
following:
{\begin{itemize}
\item \textbf{Input}: finite sets $P$ and $N$ of strings.
\item \textbf{Output}: a grammar $g$ (in the sense of Chomsky) that
  \EMPH{precisely} captures $P$ and $N$: all strings in $P$ are
  accepted by $g$, and all strings in $N$ are rejected.

\end{itemize}}
This model was first studied in detail by \cite{GoldEM:lanideitl} for
regular grammars. He showed that, \EMPH{``in
  the limit''} this problem cannot be solved from positive examples
alone (\IE $N = \emptyset$).  Such grammar inference with regular languages is simplistic as
the complex syntactic rules of natural language are not regular, but
regular languages are more mathematically tractable, and hence are a suitable
starting point.

On the machine learning side, in particular in  natural language processing (NLP), grammar inference and learning of Chomsky-style grammars and other formalisms has a long history; for surveys, we refer the reader to
\cite{d2011survey} and \cite{muralidaran2021systematic}. While they do not learn an explicit grammar, large transformer-based models have been shown to learn internal representations akin to implicit grammars \cite{tenney2019bert}.
Other 
inference tasks also have a long-standing tradition in NLP, with many
datasets and challenges proposed for Natural Language Inference
\cite{bowman2015large,williams2017broad,wang2019superglue,nie2019adversarial}.
Addressing REI with generative models also relates to a long line of research into sequence
generation tasks such as presented in \cite{colin2016webnlg,budzianowski2018multiwoz} and \cite{guan2021openmeva}.

Neural models have been successfully applied in all of these areas and related tasks.
Recent advances in deep learning (DL), especially the transformer architecture
\cite{vaswani2017attention}, have led to neural models now dominating virtually any NLP
task.  In the wake of this, \EMPH{code synthesis}, i.e.,
the task of generating programming language code based on natural
language instructions, has become an interesting next step; transformers in general, and
GPT-like models \cite{radford2019language,brown2020language} in particular are dominating
this task as well, achieving top scores on datasets
such as MBPP \cite{austin2021program} and APPS \cite{hendrycks2021measuring}.
We posit there is a direct connection between code synthesis and regular expression generation.

\PARAGRAPH{REI as a Yardstick for Quantifying ML Progress}
Grammar inference has two core quantifiable dimensions:
\begin{itemize}
\item \textbf{Correctness:} what fraction of data is classified correctly by the
  learned representations?
\item \textbf{Optimisation:} how far is the cost of a learned representation away from the achievable
  minimum?
\end{itemize}
Recently, GI was brought into focus by
\cite{DeletangG:neunetatch,vanderpoel2023mlregtest} who use benchmarks
directly related to the Chomsky hierarchy. Both observe strong
correlations between DL architectures and position of a benchmark in
the hierarchy.  What all existing GI benchmarks have in common is their  focus on 
\emph{correctness}; this leaves \emph{optimisation}, a core goal of ML on the road toward AGI, 
unquantified.

The challenge presented here changes this. To our knowledge, we
present the first benchmark to focus on quantification of a learner's
ability to \textbf{optimise while remaining correct}. We do this in a
 simple and natural way: like previous work on GI, we ask learners to produce a grammar (REs)
from positive and negative examples. Unlike previous work, we quantify how far away the learned grammar is from the
possible minimum cost. A grammar is \EMPH{optimal} if it achieves minimal cost while correctly classifying all examples.

We deliberately frame REI as a \emph{challenge} benchmark, which we introduce here. We provide training data and an evaluation harness, as well as first approaches to the task using heuristics as well as modern DL techniques, in particular, LLM-based inference. These suggest that REI is indeed a hard optimisation problem, and we suggest using our benchmark as a stepping stone for quantifying and improving the optimisation abilities of contemporary ML models.

There are multiple reasons for REI being a compelling challenge
that helps us understand and quantify important dimensions of
modern ML.

The computational complexity of this
problem is well understood: \cite{GoldMD:comoautifgd,AngluinD:onthecomlomiors,PittL:mincondpcbawap,KearnsM:crylimolbfafa} showed that
(a small variant of) this problem is NP-hard, and NP-hard even to approximate. 
They also showed that hardness does not depend on
the underlying representation of regular languages, whether DFAs, NFAs, REs, or indeed any other.  Those hardness results assure us
that there is no easy shortcut that has been overlooked
so far.  We can also easily adjust the hardness of the problem by
increasing simple parameters such as the size of input
examples. Moreover, not only are NP-hard problems extremely well
studied, but most properties of regular languages/expressions are
known, such as their expressive power and limitations, their relationships with  automata and grammar formalisms, the computational complexity of expression minimisation, equivalence
checking, and more.

Another advantage of REI is that hardness can easily be adjusted 
in other ways. For example by searching over harder classes of
languages (context-free or context-sensitive), or simpler ones, e.g., REs
with limited star-height.   Another form of simplifying REI is by
allowing the learner to query the system to be learned, leading to
Angluin-style active automata learning \cite{AngluinD:learegsfqac}, a
form of ML that is actively used in formal verification and
computer security \cite{10.1145/2967606}.  REI is also directly
related to other applications, \EG circuit complexity,  bio-informatics, and 
network security.
All those applications and connections with NP-hard problems mean that
we have at our disposal a large and heterogeneous toolset to
approach the inference challenge. Conversely, each approach that
does well on it can be adapted to help with many other
important problems.

A further advantage of the challenge is that REI is, to the best of
our knowledge, an unsolved problem for deep learning based ML.
While neural regular expression synthesis has
been investigated 
\cite{10.5555/645517.655778,locascio2016neural,ZhongZ:semregasbafgrefnls,ParkJU:sofreggrfnldusre,LiY:traregmmresbgar,chen2023data},
no existing work allows for configurable cost function, or  achieves minimality; indeed, none even guarantee that
all strings in $P$ and $N$ are classified correctly.

From the point-of-view of furthering research in transformer-based
large language models (LLMs), REIC has another
advantage: each problem instance is small, at most a few hundred
tokens. Hence, limited attention window size is irrelevant; REI
is hard already when it fits comfortably inside even small attention
windows. Therefore, it enables fine-grained and principled investigations
into LLM-learning inside the attention window, and neatly separates
the problems of quantifying the generalisation performance of
transformers inside the attention window from issues arising from the
limitations of window size.  This is especially relevant as
REI can be seen as an idealisation of code synthesis; most contemporary tools for this task are
transformer-based LLMs \cite{wang2021codet5,chen2021evaluating,chatgpt,copilot,li2023starcoder,Tunstall2023starchat-alpha}.

\section{The Regular Expression Inference Challenge}
Formally, given a finite alphabet $\Sigma$, the \EMPH{regular
expressions over $\Sigma$}, short $\RE{\Sigma}$ are given by the
grammar below, where option (?), Kleene-star (*) and complement
($\sim$) are unary operators; concatenation ($\cdot$), intersection ($\INTERSECT$), union (+), and restriction ($-$) are binary operators:
\[
r, r' ::= 
\begin{cases}

  \emptyset & \text{The empty set}\\

  \epsilon & \text{The empty string}\\

  a & \text{Alphabet character}\ a \in \Sigma\\
  
  r? & \text{Shorthand for }\epsilon+r \\
  
  r^* & \text{0 or more repetitions of }r\\
  
  \sim r & \text{Complement}\\

  r\cdot r' & \text{Concatenation}\\
  
  r \INTERSECT r' & \text{Intersection, logical conjunction}\\

  r+r' & \text{Union, logical disjunction}\\

  r-r' & \text{Restriction}

\end{cases}
\]
We use standard abbreviations, \EG $rr'$ for $r \cdot r'$. 
  With each $r \in \RE{\Sigma}$ we associate  the \EMPH{language of $r$}, 
  $\LANG{r}$:
  {\begin{itemize}
  \item[] \begin{itemize}
  \item $\LANG{\emptyset} = \emptyset$, $\LANG{\epsilon} = \{\epsilon\}$,  $\LANG{a} = \{a\}$
  \item  $\LANG{r?} = \LANG{\epsilon + r}$
  \item $\LANG{r^*} = \bigcup_{n \geq 0} \LANG{r^n}$ where $r^0 = \epsilon$ and $r^{n+1} = r \cdot r^n$
  \item $\LANG{\sim r} = \Sigma^* \setminus \LANG{r}$
  \item $\LANG{r \cdot r'} = \LANG{r} \cdot \LANG{r'}$
  \item $\LANG{r \INTERSECT r'} = \LANG{r} \cap \LANG{r'}$
  \item $\LANG{r + r'} = \LANG{r} \cup \LANG{r'}$
  \item $\LANG{r - r'} = \LANG{r} \setminus \LANG{r'}$    
  \end{itemize}
  \end{itemize}}

This induces an equality on REs: $r$ is equivalent
to $r'$ iff $\LANG{r} = \LANG{r'}$. For example, $r+r$ and $r$ have
the same language; likewise $\LANG{r^{**}} = \LANG{r^*}$.  Note that
each equivalence class has an infinite number of inhabitants.  

A \EMPH{cost function} is a map $ \COST{\cdot} : \RE{\Sigma} \FS
\NAT$ such that there are constants $c_1, \ldots, c_8 > 0$  with\footnote{We use this family of cost-functions for several reasons: (i)  existing  works on the  complexity of REI use ``uniform cost'', (ii) \cite{valizadeh2023search} uses the same family,  (iii) our cost-functions are 8-tuples of integers, hence easy to learn---it is unclear how to specify more complex cost functions, and this reassures us that REI hardness is not an effect of how the cost functions are given.}:
{\begin{itemize}
\item[]\begin{itemize}
  \item    $\COST{\emptyset} = \COST{\epsilon} = \COST{a} = c_1\ \forall a \in \Sigma$
  \item    $\COST{r?}  = \COST{r} + c_2$
  \item    $\COST{r^*} =   \COST{r} + c_3$
  \item    $\COST{\sim r}  = \COST{r} + c_4$
  \item    $\COST{r \cdot r'} = \COST{r} + \COST{r'} + c_5$
  \item    $\COST{r \INTERSECT r'} = \COST{r} + \COST{r'} + c_6$
  \item    $\COST{r + r'} = \COST{r} + \COST{r'} + c_7$
  \item    $\COST{r - r'} = \COST{r} + \COST{r'} + c_8$   
  \end{itemize}\end{itemize}}

We call each $c_i$ the \EMPH{cost} of the corresponding regular operators.
An important special case is the \EMPH{uniform cost} given by setting
$c_1 = \cdots = c_8 = 1$.
Expressions in $\RE{\Sigma}$ form tree structures defining the scope
of operators; when writing expressions as linear strings, we use parentheses
 where necessary to clarify scope,  they don't count towards the RE's cost.

Finally, regular expression inference is a supervised learning problem and has the following structure:

{\begin{itemize}
\item \textbf{Input:} Two finite sets of strings $P$ and $N$ of positive and negative examples, and a cost function $\COST{\cdot}$ for REs.
\item \textbf{Output:} A regular expression  $r$ that is both:
\begin{itemize}
  \item \textbf{Precise:} Meaning that $r$ accepts all strings in $P$ and rejects all strings in $N$.
  \item \textbf{Minimal:} No regular expression with a cost less than
     $\COST{r}$ is precise\footnote{Note there may be
other precise $r'$ with $\COST{r'}=\COST{r}$.}.
  \end{itemize}
\end{itemize}}
We call the pair $P, N$ a \EMPH{PN-set}, the combination with a cost
function an \EMPH{instance}, and $r$ the \EMPH{solution} of the
instance.

We can adjust the hardness of REI. For example, as every REI problem $P, N$ can be solved at a cost not
exceeding that of the trivial solution, \IE the union of all strings in $P$,
we can trivially rule out  operators (other than union,
concatenation and characters) by setting their cost to something above the cost of the trivial solution. We also note that we can make REI simpler, e.g., if instead of using the cost function as an input
parameter, we fix a single cost function for all instances.

\section{Datasets}
\label{sec:data}

\begin{table*}[ht!]
\center
{
{\small
\begin{tabularx}{.84\textwidth}{lccccccc}
\toprule
Dataset & $\Sigma$ & RE Characters \& Ops & Cost Function & \#PN Sets & \#Instances & \#Train & \#Test \\
\midrule
DS1 & $\{0,1\}$ & $\{\epsilon, a, ?, *, \cdot, +\}$ & uniform & $6,053$ & $6,053$ & $5,447$ & $606$\\
DS2 & $\{0,1\}$ & $\{\epsilon, a, ?, *, \cdot, +\}$& variable & $6,029$ & $120,580$ & $110,329$ & $10,251$\\
DS3 & $\{0,1\}$ & $\{\emptyset, \epsilon, a, ?, *, \sim, \cdot, \&, +, -\}$ & uniform & $6,054$ & $6,054$ & $5,448$ & $606$\\
DS4 & $\{0,1\}$ & $\{\emptyset, \epsilon, a, ?, *, \sim, \cdot, \&, +, -\}$ & variable & $6,013$ & $120,260$ & $108,234$ & $12,026$\\
\bottomrule
\end{tabularx}}
}
\caption{The four datasets generated for the REI challenge. Here and elsewhere \#S denotes the cardinality of $S$.}
\label{tab:data}
\end{table*}

We now detail the data released as part of the challenge. The data is
produced automatically using an extension of the  GPU-accelerated REI solver of
\cite{valizadeh2023search}: we added
three additional operators ($\sim, \&, -$). This allowed us to introduce datasets~DS3 and DS4, as explained below.
The data is split into official training and test
sets; the former, but not the latter, come with associated solutions,
\IE for each PN-set and cost function, a minimal regular expression.

Using an algorithmic solver to generate the data for the machine learning challenge guarantees that it is correct: the solver will output an expression for a given PN-set and cost function \emph{if and only if} it is indeed minimal \WRT that input\footnote{PN-set and cost instances for which the solver fails are not included in the released data.}.

We provide four separate datasets, differing in key characteristics along
two dimensions:
{\begin{itemize}

  \item \textbf{Operators.} Datasets~1 and 2 (DS1, DS2) allow only the RE operators 
    for option ($?$), Kleene-star ($*$), concatenation ($\cdot$), and union
    ($+$). This means complement ($\sim$), intersection
    ($\INTERSECT$) and restriction $(-)$ are not available. Datasets~3 and 4 (DS3, DS4) allow all  operators.

    \item \textbf{Cost functions.} We
    use only the uniform cost for Datasets~1 and 3, while making
    it variable for Datasets~2 and 4, which provide a number of
    solutions per PN-set, given different random cost functions.

\end{itemize}}
We posit that such a division makes the problems in REIC
more variable; we explicitly encourage participants to contribute to
results on only some,  or all of the four datasets.

When splitting the four generated datasets into training and test
data, we aim for a $90/10$ split. For each of the datasets, we ensure
that the test portion only contains regular expressions that \emph{do
not occur} in the training part. However, sometimes solutions to
a PN-set in the training set might also be solutions to an instance
in the test set.

\subsection{Data Generation}\label{subsec:datasetsGeneration}
Our data generation approach has two parts, random PN-set and random cost function generation.

\PARAGRAPH{PN-set Generation}
Random PN-sets are generated with a small set of parameters:

{\begin{itemize}

\item $\Sigma$: The set of characters used in $P$ and $N$.

\item $le$: The maximum length of example strings in $P$ and $N$ (0 is always minimum length).

\item $p$, $n$: number of positive negative examples respectively,
\IE size of $P$ and $N$.
\end{itemize}}
For these parameters, we use two PN-generation schemes from strings
over $\Sigma$, with complementary properties.
{\begin{itemize}
	
\item \textsc{Type 1}:
      We uniformly and independently sample $p$ strings from the set
      of all strings of size $0, 1, ..., le$ , and $n$
      strings for $N$.  \textsc{Type 1} is heavily biased towards long
      strings, since there are exponentially more long strings than
      short ones. Since short strings, in particular $\epsilon$ and the
      characters in $\Sigma$, play an important role in regular
      languages, bias towards long strings is sometimes sub-optimal,
      and we use a complementary second scheme.
  
\item \textsc{Type 2}:
      This scheme gives more weight to shorter
      strings and works in two steps: first, uniformly and
      independently sample a target string length $i$, and then
      uniformly and independently sample a string of length $i$.
      We use this procedure to generate strings for P and N until the
      sizes $p$ and $n$ are reached.

\end{itemize}}
\noindent In both cases, we ensure that $P$ and $N$ are disjoint, to rule out
challenge problems without solutions.
As an illustration, assuming an alphabet $\Sigma=\{0, 1\}$, an example of a generated
Type 1 PN-set with parameters $le=~6, p=12, n=10$ is
{\small{
\begin{align*}
\text{P:} &\quad 10110, \ 001110, \ 010111, \ 1, \ 00101, \ 011111, \ 011,\\
 & \quad 000000, \ 0111, \ 10011, \ 001100, \ 001001 \\
\text{N:} & \quad 000, \ 00010, \ 010010, \ 010011, \ 01010, \ 100001, \ 1011,\\
& \quad 110000, \ 11001, \ 1110
\end{align*}
}}

\noindent With the same alphabet, Type 2 generation with parameters $le=8, p=10, n=7$ produces, for example
{\small{
\begin{align*}
\text{P:} & \quad 000,\ 0101,\ 1110010,\ 11,\ 1,\ 111,\ \epsilon ,\ 11101100,\ 0, 11001\\
\text{N:} & \quad 00010,\ 0010011,\ 010,\ 01001000,\ 011,\ 10,\ 11010
\end{align*}
}}

The four datasets discussed above contain instances generated (roughly) evenly with \textsc{Type~1} and \textsc{Type~2} over the alphabet $\{0,
1\}$. For \textsc{Type~1}, parameters $p$ and $n$ range over all of $1,\ldots,10$, and $le$ ranges over all of $0,\ldots,7$;
for \textsc{Type~2}, we use the same ranges for $p$ and $n$, and $le$ from $0$ to $10$.

For the challenge presented here, we restrict ourselves to strings over a binary alphabet for two reasons:
First, as our baselines in the next section indicate clearly, REI with a binary alphabet is
already hard to learn. Secondly, getting ground truth for substantially
bigger alphabets, \EG ASCII, is currently infeasible, because the performance even of the SOTA GPU-based REI solver we use \cite{valizadeh2023search} is heavily constrained by alphabet size.

\PARAGRAPH{Cost Function Generation}
As mentioned above, our datasets are split \WRT cost functions: In DS~1 and 3 the cost
function is fixed to the uniform cost.  In the remaining DS~2 and 4 the cost function is
a parameter. To generate costs, we randomly and independently sample from $1, \ldots, 49$ for each
$c_i$ of the cost function. For DS~4, we use all $c_i$; since DS~2
uses a reduced set of operators, we only sample costs of relevant operators.
For both datasets, we generate 19 random cost functions in addition to the uniform cost,
for each PN-set. Hence $k$ PN-sets give us $20*k$ inference problems.
Table~\ref{tab:data} shows an overview of the data generated for REIC.

After generating PN-sets and cost functions, we use the REI solver from \cite{valizadeh2023search} to generate reference minimal REs for each instance.


\section{REIC Metrics and Baselines}
\label{sec:exp}

\paragraph{Challenge scoring.}
The REI challenge is scored using the following metrics.  All scoring
is automated.  All scores below are calculated for all instances in a dataset; if
an instance solution is not syntactically valid, it will also negatively
contribute to other scores.
{\begin{itemize}

\item \textbf{Compile Ratio}: The ratio of generated REs that are syntactically valid, i.e., that can be
  parsed by a RE parser. 

\item \textbf{Precise Absolute}: Number of instances for which the generated RE correctly classifies the entire PN-set.

\item \textbf{Precise Ratio}: The ratio of instance REs that correctly
  accept and respectively reject the PN-sets.

\item \textbf{Positive Ratio}: The ratio of positive examples that are correctly
accepted.

\item \textbf{Negative Ratio}: The ratio of negative examples that are correctly
rejected.

\item \textbf{PN Ratio}: The ratio of combined accepted strings in P with
  rejected strings in N.

\item \textbf{Minimal Instances}: The number of instances for which
  the generated RE is precise and minimal.

\item \textbf{Minimal Ratio -- Precise}: The ratio of minimal REs over
  precise generated REs.

\item \textbf{Minimal Ratio -- Global}: The ratio of minimal REs over
  all test instances.

\item \textbf{Cost Ratio}: The average ratio of precise generated REs'
  costs over the gold-standard minimal RE costs.

\end{itemize}}
Our leaderboard lists all scores for participating
models\footnote{As heuristics/algorithms are known to achieve perfect scores, the official leaderboard only lists and ranks ML/DL models.}. The leader is determined solely by global
Minimal Ratio\footnote{`Gaming''  the system to
get high scores in some metrics is easy; e.g., the Trivial
baseline achieves 100\% Compile Rate, Precision, and P/N/PN Ratio by
definition. Similarly, Minimal Ratio (Precise) and Cost Ratio can easily be exploited by high-precision, low-recall systems. Hence, any score encompassing a (weighted) average score could easily be artificially tuned.}.

\paragraph{Baselines.}
We devise REI as a challenge to entice the machine learning/deep learning community to explore this hard task. Addressing REI with ML/DL methods is as of yet an unsolved problem; while perfect algorithmic solvers exist (in fact, one such solver was used to generate the data for REIC), no such perfect learned model is known.

To bootstrap REIC with baseline solutions, and provide further insights into the challenge, we devise several heuristic baselines, as well as two first DL-based approaches:

{\begin{itemize}

\item The \textbf{Trivial RE} baseline simply generates, for each
  test instance, the trivial regular expression formed by
  the union ($+$) of the strings in its positive set. Note that this
  always returns a precise RE, albeit usually not minimal.

\item \textbf{PN Retrieval} is a baseline that, for each test instance, retrieves 
the closest matching train data instance \WRT overlap of their PN sets, and uses
the corresponding regular expression for the test. If
multiple PN sets the with same overlap are retrieved, the lowest cost RE
\WRT the test instance cost function is returned. This baseline lets use assess
how much of the training data is solvable by learning from closely related data.

\item \textbf{RE Retrieval} has access to all regular expressions in given training sets (DS1+DS2, DS3+DS4),
and tests all of them against each test data instance, selecting
for each that RE with highest PN Ratio. In cases of ties, it selects that expression with lowest cost, according
to the test instance cost function. This baseline can show us how many test instances are solvable
precisely and with minimal cost with expressions seen in the training data.

\item \textbf{StarChat$\pmb{\beta}$} uses a pretrained code LLM and performs REI with a few-shot prompt. Specifically, we employ \emph{StarChat$\beta$} \cite{Tunstall2023starchat-alpha}, an instruction-tuned variant of StarCoderPlus \cite{li2023starcoder}, a SOTA $15.5$~billion parameter code LLM. We use prompting to instill information about REI into the model, and let it generate 10 regular expressions for each test instance, keeping the best. Prompts and inference details used for this baseline are shown in the technical appendix.

\item \textbf{ReGPT} is a GPT-2-like \cite{radford2019language} model,
  trained on the training portions of the REIC datasets; details are given below, and all training hyperparameters are given in the technical appendix.
  We use the model's inference capabilities as a
  generative LLM to generate regular expressions for the test set instances.

\end{itemize}}

\begin{table*}[ht!]
\center
{
{\small
\begin{tabularx}{.81\textwidth}{l|cccccccc|c|c}
\toprule
 & CR & Prec & Prec\% & P\% & N\% & PN\% & Min & Min\% P& \textbf{Min\% G} & Cost Ratio \\
\midrule
& \multicolumn{10}{c}{Dataset 1}\\
Trivial & 100 & 606 & 100 & 100 & 100 & 100 & 0 & 0.0 & 0.0 & 3.34\\
PN Ret & 100 & 3 & 0.5 & 45.9 & 89.3 & 67.8 & 1 & 33.3 & 1.7 & 1.20\\
RE Ret & 100 & 302 & 49.8 & 94.0 & 95.6 & 95.4 & 133 & 44.0 & 21.9 & 1.15\\
StarChat$\beta$ & 100 & 5 & 0.8 & 37.2 & 85.9 & 63.4 & 0 & 0.0 & 0.0 & 2.02\\
ReGPT-1k & 100 & 60 & 9.9 & 80.9 & 88.5 & 87.4 & 33 & 55.0 & 5.4 & 1.10\\
ReGPT-10 & 100 & 10 & 1.7 & 65.6 & 74.2 & 73.4 & 7 & 70.0 & 1.2 & 1.08\\
\midrule
& \multicolumn{10}{c}{Dataset 2}\\
Trivial & 100 & 10251 & 100 & 100 & 100 & 100 & 77 & 0.8 & 0.8 & 3.35 \\
PN Ret & 100 & 10251 & 100 & 100 & 100 & 100 & 2133 & 20.8 & 20.8 & 1.08\\
RE Ret & 100 & 10251 & 100 & 100 & 100 & 100 & 2375 & 23.2 & 23.2 & 1.06\\ 
StarChat$\beta$ & 99.69 & 29 & 0.3 & 39.6 & 79.8 & 61.9 & 3 & 10.3 & $<$0.1 & 1.47\\
ReGPT-1k & 100 & 9787 & 95.5 & 99.1 & 99.1 & 99.2 & 2013 & 20.6 & 19.6 & 1.08\\
ReGPT-10 & 99.99 & 9775 & 95.4 & 99.0 & 99.1 & 99.1 & 1945 & 19.9 & 19.0 & 1.08\\
\midrule
& \multicolumn{10}{c}{Dataset 3}\\
Trivial & 100 & 606 & 100 & 100 & 100 & 100 & 2 & 0.3 & 0.3 & 3.99\\
PN Ret & 100 & 9 & 1.5 & 63.6 & 72.5 & 70.8 & 3 & 33.3 & 5.0 & 1.20\\
RE Ret & 100 & 328 & 54.1 & 95.7 & 94.6 & 96.0 & 151 & 46.0 & 24.9 & 1.16\\ 
StarChat$\beta$ & 100 & 6 & 1.0 & 44.6 & 79.8 & 65.8 & 1 & 16.7 & 0.2 & 2.10\\
ReGPT-1k & 100 & 97 & 16.0 & 85.5 & 89.6 & 89.8 & 41 & 42.3 & 6.8 & 1.14\\
ReGPT-10 & 100 & 28 & 4.6 & 73.3 & 77.0 & 79.0 & 15 & 53.6 & 2.5 & 1.11\\
\midrule
& \multicolumn{10}{c}{Dataset 4}\\
Trivial & 100 & 12026 & 100 & 100 & 100 & 100 & 32 & 0.3 & 0.3 & 4.32 \\
PN Ret & 100 & 12026 & 100 & 100 & 100 & 100 & 1239 & 10.3 & 10.3 & 1.08\\
RE Ret & 100 & 12026 & 100 & 100 & 100 & 100 & 1412 & 11.7 & 11.7 & 1.07\\ 
StarChat$\beta$ & 100 & 52 & 0.4 & 44.5 & 78.9 & 64.2 & 1 & 1.39 & $<$0.1 & 2.56\\
ReGPT-1k & 100 & 10222 & 84.99 & 97.0 & 97.2 & 97.4  & 1093 & 10.7 & 9.1 & 1.09\\
ReGPT-10 & 99.99 & 9590 & 79.7 & 94.7 & 95.0 & 95.3 & 957 & 10.0 & 8.0 & 1.09\\
\midrule
\midrule
& \multicolumn{10}{c}{All Data}\\
Trivial & 100 & 23489 & 100 & 100 & 100 & 100 & 111 & 0.5 & 0.5 & 3.86 \\
PN Ret & 100 & 22289 & 94.9 & 97.7 & 99.0 & 98.4 & 3376 & 15.1 & 14.4 & 1.08\\
RE Ret & 100 & 22907 & 97.5 & 99.7 & 99.7 & 99.8 & 4071 & 17.8 & 17.3 & 1.07\\ 
StarChat$\beta$ & 99.86 & 92 & 0.4 & 42.1 & 79.5 & 63.2 & 5 & 5.4 & $<$0.1 & 2.16 \\
ReGPT-1k & 100 & 20166 & 85.9 & 97.2 & 97.6 & 97.7  & 3180 & 15.8 & 13.5 & 1.09\\
ReGPT-10 & 99.99 & 19403 & 82.6 & 95.3 & 95.8 & 96.0 & 2924 & 15.1 & 12.4 & 1.09\\
\bottomrule
\end{tabularx}}
}
\caption{Baseline results on the four REIC test datasets, as well as combined test data. PN Ret and RE Ret are PN Retrieval and RE Retrieval baselines, respectively. CR refers to Compile Ratio, Prec/Prec\% to total/ratio of REs that precisely accept/reject PN-sets, P/N/PN\% to accuracies over respective sets, Min\% P/Min\% G to ratio of minimal cost for precise REs/across full dataset. ``ReGPT-1k'' and ``ReGPT-10'' are our ReGPT model with $1k$ and $10$ sampled solutions, respectively.}
\label{tab:results}
\end{table*}

We considered \emph{automaton minimisation} as a baseline, such as taking an NFA $\mathcal{A}$, \EG one based on the trivial solution, and transforming it into an equivalent minimal DFA $\mathcal{B}$. However, we decided against it for several reasons: First, by definition, $\mathcal{B}$  accepts the same language as $\mathcal{A}$; but this is not necessarily the case for a minimal solution to a REI problem -- it just needs to classify the PN-set correctly, regardless of the language it specifies. Hence $\mathcal{B}$ will typically not be minimal for a given PN-set. Second, to the best of our knowledge, no algorithm for DFA minimisation exists that can handle cost functions. Finally, even assuming a notion of minimality in DFAs, \EG number of states, we doubt the standard translations from automata (possibly cyclic graphs) to REs (trees) preserve minimality. We therefore disregard this as a baseline suited to the challenge at hand.

\paragraph{ReGPT Training and Inference.}
For ReGPT we train two models from scratch, one on the combined training sets DS1 and DS2, and a second model on DS3 and DS4. All models are implemented in the HuggingFace
transformers
framework\footnote{\url{https://huggingface.co/docs/transformers}}
and use the GPT-2 architecture with a total of ${\sim}300$M parameters. Details are shown in the technical appendix.

We use a \texttt{[CLS]} token to indicate the beginning of the
input, \texttt{[POS]} and \texttt{[NEG]} for
positive/negative strings, one \texttt{[COST\_X]} token per operator to indicate the cost function, and \texttt{[BOR]} and
\texttt{[EOR]} to mark the beginning/end of a regular expression;
\texttt{[EOR]} doubles as the end-of-sequence token. An example ReGPT encoding is shown in the appendix.

During training, we randomly split the combined training data into train and validation sets in a 90/10 split. To select the model checkpoints for final evaluation on the test sets, we greedily generate one solution for each validation instance, and calculate the average of PN Ratio and Global Minimality. For each test set, we select the checkpoint with highest average score on the respective validation set.

For inference on the test set, we use the trained ReGPT models to generate $1k$ solutions for each instance: one greedily, and the remaining solutions using Nucleus Sampling \cite{holtzman2019curious}.
We keep the best solution as measured by PN Ratio; in cases of ties, we keep the solution with lower cost according to the test instance cost function.

In our inference setup, sampling $1,000$ REs with ReGPT takes roughly the same time as sampling $10$ solutions with StarChat$\beta$. However, for a better comparison, we also report ReGPT performance when generating 10 solutions. We provide training and inference details in the appendix.

Note that these first baselines are in fact not directly optimized for RE minimalisation. We provide these as examples of addressing the novel challenge we present with the currently dominant approach to inference tasks, i.e., training LLMs on large amounts of positive data. While the cost function is encoded in the LLM input, there is no theoretical guarantee that the model will be able to generalise to unseen combinations of PN-sets and cost functions. This is deliberate: We provide a first mainstream approach to the new challenge benchmark, and leave more involved approaches, e.g., Reinforcement Learning using cost functions for the reward signal, to future work and future challenge entries.

\subsection{Discussion of Baseline Performance}
Table~\ref{tab:results} summarises the performance of all our baselines on the four test datasets, as well as on all combined test data.

The heuristic baselines -- \emph{Trivial RE, PN Retrieval, and RE Retrieval} -- provide valuable insights into the challenge data.
Of course, the Trivial baseline achieves 100\% in all PN coverage metrics, as well as Compile Rate. Due to data artifacts, it even ``finds'' some minimal expressions; however, these finds are rare, showing that the minimal RE is almost never trivial. Across all datasets, this baseline achieves the worst Cost Ratio, producing expressions up to more than $4.3$ times as costly on average than the minimal regular expression, on Dataset 4.

The two retrieval baselines reveal an interesting pattern across the four datasets. While both PN and RE Retrieval methods are able to find expressions that are $100\%$ \emph{precise} on DS2 and DS4, this metric falls to ${\sim}50\%$ (RE) and only ${\sim}1\%$ (PN) on the (much smaller) DS1 and DS3. For minimality, retrieving solutions based on PN sets or REs leads to very similar performance on the datasets with variable cost functions, while there is a stark performance difference on those with uniform cost function. This is likely an artefact of data generation, where $20$ instances per PN set are given for both DS2 and DS4, with varying cost functions. Therefore, the retrieval baselines are able to find corresponding instances across these larger training and test data; while this leads to perfect precision, the heuristics also reveal that at most $1/4$ (RE Retrieval, $24.9\%$ minimality on DS3) of the challenge is solvable from the data directly, without learning techniques.

ReGPT (1k) produces syntactically valid regular expressions for all test instances, indicating it has learned a good internal representation of the RE grammar. Even when only sampling 10 solutions, it performs at or essentially at 100\% compile rate. While it is outperformed by the (very strong) RE Retrieval, ReGPT generates better expressions than can be retrieved by PN Retrieval on the smaller test sets, DS1 and DS3.
PN Retrieval is conceptually the closest heuristic to the learned model, which is trained to generate REs conditioned on PN sets; ReGPT outperforming this baseline on the smaller datasets -- which are more out-of-distribution compared to the training data than the larger test sets -- indicates that there is a learning signal available in the training data that allows a degree of generalisation.
ReGPT's strongest test sets are DS2, followed by DS 4, which both contain variable cost functions for the same PN sets.
These datasets of course contribute the vast amount of training data during ReGPT training; in addition, Dataset 2 uses the reduced set of permitted operators, explaining the relatively high test performance.

Across all datasets, ReGPT finds cost-efficient solutions; on the challenging sets DS1 and DS3, the average cost of its perfect solutions even is \emph{lowest} across systems. This yields interesting possibilities of using trained models in combination with algorithmic solvers such as that of \cite{valizadeh2023search}, for example by providing precise solutions that are \emph{close} to minimal as starting points for the search algorithm, potentially helping with search space reduction.

The StarChat$\beta$ baseline shows a severe performance gap on the challenge test sets. This is not  surprising: While it is truly large and trained for code synthesis -- a task closely related to, if not a superset of regular expression inference -- it has never seen REIC data during training. While prompting instruction-tuned LLMs with specific tasks can be a powerful technique, it seems not enough to perform well on the novel, untrained, and provably hard challenge proposed here.

In fact, as StarChat$\beta$'s compile rate of (essentially) 100\%  demonstrates, the model is capable of generating REs as defined for our challenge. However, when it comes to \emph{precision} and \emph{minimality}, the large SOTA pretrained code LLM struggles with the proposed challenge in a prompt-based setting, where it underperforms Compared to the much smaller, but fully supervised ReGPT baseline. This is true even when comparing models that sample the same number of outputs.

It is possible (in fact likely) that a fine-tuned version, i.e., a StarChat$\beta$ model trained on REIC's training data, will outperform the prompt approach -- though competitiveness with ReGPT remains another question\footnote{We speculate that, given the small context size for REI, a fully supervised model like ReGPT should not require billions of parameters to properly capture the task.}. However, training such a large model even for finetuning is   time-consuming and expensive. Other prompting methods -- either with different hand-written prompts to ours, or trained -- might also yield improvement; however, the long inference time and resources needed
make rapid development of this a challenge in itself. As this work aims to establish Regular Expression Inference as a challenge for the research community, we thus leave these approaches as interesting future work. 

Finally, focusing on model performance on all combined data, it seems clear that the hardness of REIC indeed lies \emph{in the task of finding minimal solutions} to the given inference problems. The heuristic baselines perform close to $100\%$ precision, and ReGPT's learning approaches $86\%$ for the combined data on the same metric, which we can expect to be improved upon with more sophisticated models. With exception of StarChat$\beta$, all approaches including ReGPT achieve scores in the high $90$s on \emph{PN Ratio}, giving further evidence that learning to cover the PN-sets is achievable. However, even the strongest heuristic baseline, RE Retrieval, only finds precise and \emph{minimal} solutions on $17.3\%$ of the full data.

It seems clear that REI provides a challenging setting, for both heuristic and learning-based methods. We hope these results will incentivise the research community to actively participate in the REIC, pushing the envelope in regular expression and deep learning research and understanding.

\section{Conclusions}\label{conclusion}

We have introduced the regular expression inference challenge, an
open-ended research challenge with a novel goal, \emph{optimisation while remaining correct}, for the machine learning community, inviting participants to develop models that can generate minimal regular expressions
\WRT PN-sets and cost functions.
We motivated the challenge in the light of regular expression and formal language research, machine learning research, and potential practical applications that we believe stand to benefit from regular expression inference.
We employed heuristic and learned baselines, indicating that the REI challenge is difficult, and warrants further research into how machine learning and deep learning approaches can be leveraged to better cover this problem.

We invite the research community to participate in the REI challenge hosted on CodaLab, where starter code is also made available to enable reproduction of our baselines:

\begin{center}
\url{https://codalab.lisn.upsaclay.fr/competitions/15096}
\end{center}
\clearpage

\bibliographystyle{named}
\bibliography{ijcai24}

\begin{thebibliography}{}

\bibitem[\protect\citeauthoryear{Angluin}{1978}]{AngluinD:onthecomlomiors}
Dana Angluin.
\newblock On the complexity of minimum inference of regular sets.
\newblock {\em Information and Control}, 39(3):337--350, 1978.

\bibitem[\protect\citeauthoryear{Angluin}{1987}]{AngluinD:learegsfqac}
Dana Angluin.
\newblock {Learning Regular Sets from Queries and Counterexamples}.
\newblock {\em Inf. Comput.}, 75(2):87–106, November 1987.

\bibitem[\protect\citeauthoryear{Austin \bgroup \em et al.\egroup }{2021}]{austin2021program}
Jacob Austin, Augustus Odena, Maxwell Nye, Maarten Bosma, Henryk Michalewski, David Dohan, Ellen Jiang, Carrie Cai, Michael Terry, Quoc Le, et~al.
\newblock Program synthesis with large language models.
\newblock {\em arXiv preprint arXiv:2108.07732}, 2021.

\bibitem[\protect\citeauthoryear{Bowman \bgroup \em et al.\egroup }{2015}]{bowman2015large}
Samuel~R. Bowman, Gabor Angeli, Christopher Potts, and Christopher~D. Manning.
\newblock A large annotated corpus for learning natural language inference.
\newblock In {\em Proceedings of EMNLP 2015}, pages 632--642, Lisbon, Portugal, September 2015.

\bibitem[\protect\citeauthoryear{Brown \bgroup \em et al.\egroup }{2020}]{brown2020language}
Tom Brown, Benjamin Mann, Nick Ryder, Melanie Subbiah, Jared~D Kaplan, Prafulla Dhariwal, Arvind Neelakantan, Pranav Shyam, Girish Sastry, Amanda Askell, et~al.
\newblock Language models are few-shot learners.
\newblock {\em Advances in neural information processing systems}, 33:1877--1901, 2020.

\bibitem[\protect\citeauthoryear{Budzianowski \bgroup \em et al.\egroup }{2018}]{budzianowski2018multiwoz}
Pawe{\l} Budzianowski, Tsung-Hsien Wen, Bo-Hsiang Tseng, I{\~n}igo Casanueva, Stefan Ultes, Osman Ramadan, and Milica Ga{\v{s}}i{\'c}.
\newblock {M}ulti{WOZ} - a large-scale multi-domain {W}izard-of-{O}z dataset for task-oriented dialogue modelling.
\newblock In {\em Proceedings of EMNLP 2018}, pages 5016--5026, Brussels, Belgium, October-November 2018.

\bibitem[\protect\citeauthoryear{Chen \bgroup \em et al.\egroup }{2021}]{chen2021evaluating}
Mark Chen, Jerry Tworek, Heewoo Jun, Qiming Yuan, Henrique Ponde de~Oliveira Pinto, Jared Kaplan, Harri Edwards, Yuri Burda, Nicholas Joseph, Greg Brockman, et~al.
\newblock Evaluating large language models trained on code.
\newblock {\em arXiv preprint arXiv:2107.03374}, 2021.

\bibitem[\protect\citeauthoryear{Chen \bgroup \em et al.\egroup }{2023}]{chen2023data}
Qiaochu Chen, Arko Banerjee, \c{C}a\u{g}atay Demiralp, Greg Durrett, and I\c{s}\i{}l Dillig.
\newblock Data extraction via semantic regular expression synthesis.
\newblock {\em Proc. ACM Program. Lang.}, 7(OOPSLA2), oct 2023.

\bibitem[\protect\citeauthoryear{Chomsky}{1956}]{chomsky1956three}
Noam Chomsky.
\newblock Three models for the description of language.
\newblock {\em IRE Transactions on information theory}, 2(3):113--124, 1956.

\bibitem[\protect\citeauthoryear{Colin \bgroup \em et al.\egroup }{2016}]{colin2016webnlg}
Emilie Colin, Claire Gardent, Yassine M’rabet, Shashi Narayan, and Laura Perez-Beltrachini.
\newblock {The WebNLG Challenge: Generating Text from DBPedia Data}.
\newblock In {\em Proceedings of the 9th international natural language generation conference}, pages 163--167, 2016.

\bibitem[\protect\citeauthoryear{Courcelle and Engelfriet}{2012}]{10.5555/2414243}
Professor~Bruno Courcelle and Dr~Joost Engelfriet.
\newblock {\em Graph Structure and Monadic Second-Order Logic: A Language-Theoretic Approach}.
\newblock Cambridge University Press, USA, 1st edition, 2012.

\bibitem[\protect\citeauthoryear{Del{\'{e}}tang \bgroup \em et al.\egroup }{2023}]{DeletangG:neunetatch}
Gr{\'{e}}goire Del{\'{e}}tang, Anian Ruoss, Jordi Grau{-}Moya, Tim Genewein, Li~Kevin Wenliang, Elliot Catt, Chris Cundy, Marcus Hutter, Shane Legg, Joel Veness, and Pedro~A. Ortega.
\newblock Neural networks and the chomsky hierarchy.
\newblock In {\em 11th International Conference on Learning Representations}, 2023.

\bibitem[\protect\citeauthoryear{D’Ulizia \bgroup \em et al.\egroup }{2011}]{d2011survey}
Arianna D’Ulizia, Fernando Ferri, and Patrizia Grifoni.
\newblock A survey of grammatical inference methods for natural language learning.
\newblock {\em Artificial Intelligence Review}, 36:1--27, 2011.

\bibitem[\protect\citeauthoryear{Firoiu \bgroup \em et al.\egroup }{1998}]{10.5555/645517.655778}
Laura Firoiu, Tim Oates, and Paul~R. Cohen.
\newblock {Learning Deterministic Finite Automaton with a Recurrent Neural Network}.
\newblock In {\em Proceedings of the ICGI 1998}, ICGI '98, page 90–101, Berlin, Heidelberg, 1998. Springer-Verlag.

\bibitem[\protect\citeauthoryear{Github}{2022}]{copilot}
Github.
\newblock {Your AI pair programmer}.
\newblock \url{https://github.com/features/copilot}, 2022.
\newblock Blog post accessed 8 June 2023.

\bibitem[\protect\citeauthoryear{Gold}{1967}]{GoldEM:lanideitl}
E.~Mark Gold.
\newblock {Language identification in the limit}.
\newblock {\em Information and Control}, 10(5):447--474, 1967.

\bibitem[\protect\citeauthoryear{Gold}{1978}]{GoldMD:comoautifgd}
E.~Mark Gold.
\newblock {Complexity of automaton identification from given data}.
\newblock {\em Information and Control}, 37(3):302--320, 1978.

\bibitem[\protect\citeauthoryear{Guan \bgroup \em et al.\egroup }{2021}]{guan2021openmeva}
Jian Guan, Zhexin Zhang, Zhuoer Feng, Zitao Liu, Wenbiao Ding, Xiaoxi Mao, Changjie Fan, and Minlie Huang.
\newblock {O}pen{MEVA}: A benchmark for evaluating open-ended story generation metrics.
\newblock In {\em Proceedings of the 59th Annual Meeting of the Association for Computational Linguistics and the 11th International Joint Conference on Natural Language Processing (Volume 1: Long Papers)}, pages 6394--6407, Online, August 2021.

\bibitem[\protect\citeauthoryear{Hendrycks \bgroup \em et al.\egroup }{2021}]{hendrycks2021measuring}
Dan Hendrycks, Steven Basart, Saurav Kadavath, Mantas Mazeika, Akul Arora, Ethan Guo, Collin Burns, Samir Puranik, Horace He, Dawn Song, and Jacob Steinhardt.
\newblock Measuring coding challenge competence with apps.
\newblock {\em NeurIPS}, 2021.

\bibitem[\protect\citeauthoryear{Holtzman \bgroup \em et al.\egroup }{2019}]{holtzman2019curious}
Ari Holtzman, Jan Buys, Li~Du, Maxwell Forbes, and Yejin Choi.
\newblock The curious case of neural text degeneration.
\newblock In {\em International Conference on Learning Representations}, 2019.

\bibitem[\protect\citeauthoryear{Hopcroft \bgroup \em et al.\egroup }{2006}]{HopcroftJE:intauttlac}
John~E. Hopcroft, Rajeev Motwani, and Jeffrey~D. Ullman.
\newblock {\em {Introduction to Automata Theory, Languages, and Computation}}.
\newblock Addison-Wesley, 2006.

\bibitem[\protect\citeauthoryear{Kearns and Valiant}{1994}]{KearnsM:crylimolbfafa}
Michael Kearns and Leslie Valiant.
\newblock {Cryptographic Limitations on Learning Boolean Formulae and Finite Automata}.
\newblock {\em J. ACM}, 41(1):67–95, jan 1994.

\bibitem[\protect\citeauthoryear{Kleene}{1956}]{KleeneSC:repeveinnafa}
S.~C. Kleene.
\newblock Representation of events in nerve nets and finite automata.
\newblock In {\em Automata Studies}, pages 3--41. Princeton University Press, Princeton, NJ, 1956.

\bibitem[\protect\citeauthoryear{Li \bgroup \em et al.\egroup }{2021}]{LiY:traregmmresbgar}
Yeting Li, Shuaimin Li, Zhiwu Xu, Jialun Cao, Zixuan Chen, Yun Hu, Haiming Chen, and Shing-Chi Cheung.
\newblock {TransRegex: Multi-Modal Regular Expression Synthesis by Generate-and-Repair}.
\newblock In {\em Proceedings of the 43rd International Conference on Software Engineering}, ICSE '21, page 1210–1222. IEEE Press, 2021.

\bibitem[\protect\citeauthoryear{Li \bgroup \em et al.\egroup }{2023}]{li2023starcoder}
Raymond Li, Loubna~Ben Allal, Yangtian Zi, Niklas Muennighoff, Denis Kocetkov, Chenghao Mou, Marc Marone, Christopher Akiki, Jia Li, Jenny Chim, Qian Liu, Evgenii Zheltonozhskii, Terry~Yue Zhuo, Thomas Wang, Olivier Dehaene, Mishig Davaadorj, Joel Lamy-Poirier, João Monteiro, Oleh Shliazhko, Nicolas Gontier, Nicholas Meade, Armel Zebaze, Ming-Ho Yee, Logesh~Kumar Umapathi, Jian Zhu, Benjamin Lipkin, Muhtasham Oblokulov, Zhiruo Wang, Rudra Murthy, Jason Stillerman, Siva~Sankalp Patel, Dmitry Abulkhanov, Marco Zocca, Manan Dey, Zhihan Zhang, Nour Fahmy, Urvashi Bhattacharyya, Wenhao Yu, Swayam Singh, Sasha Luccioni, Paulo Villegas, Maxim Kunakov, Fedor Zhdanov, Manuel Romero, Tony Lee, Nadav Timor, Jennifer Ding, Claire Schlesinger, Hailey Schoelkopf, Jan Ebert, Tri Dao, Mayank Mishra, Alex Gu, Jennifer Robinson, Carolyn~Jane Anderson, Brendan Dolan-Gavitt, Danish Contractor, Siva Reddy, Daniel Fried, Dzmitry Bahdanau, Yacine Jernite, Carlos~Muñoz Ferrandis, Sean Hughes, Thomas Wolf, Arjun Guha, Leandro von
  Werra, and Harm de~Vries.
\newblock Starcoder: may the source be with you!
\newblock {\em Transactions on Machine Learning Research}, 2023.

\bibitem[\protect\citeauthoryear{Locascio \bgroup \em et al.\egroup }{2016}]{locascio2016neural}
Nicholas Locascio, Karthik Narasimhan, Eduardo DeLeon, Nate Kushman, and Regina Barzilay.
\newblock Neural generation of regular expressions from natural language with minimal domain knowledge.
\newblock In {\em Proceedings of EMNLP 2016}, pages 1918--1923, Austin, Texas, November 2016.

\bibitem[\protect\citeauthoryear{Muralidaran \bgroup \em et al.\egroup }{2021}]{muralidaran2021systematic}
Vigneshwaran Muralidaran, Irena Spasi{\'c}, and Dawn Knight.
\newblock A systematic review of unsupervised approaches to grammar induction.
\newblock {\em Natural Language Engineering}, 27(6):647--689, 2021.

\bibitem[\protect\citeauthoryear{Nie \bgroup \em et al.\egroup }{2020}]{nie2019adversarial}
Yixin Nie, Adina Williams, Emily Dinan, Mohit Bansal, Jason Weston, and Douwe Kiela.
\newblock Adversarial {NLI}: A new benchmark for natural language understanding.
\newblock In {\em Proceedings of the 58th Annual Meeting of the Association for Computational Linguistics}, pages 4885--4901, Online, July 2020.

\bibitem[\protect\citeauthoryear{OpenAI}{2022}]{chatgpt}
OpenAI.
\newblock {Introducing ChatGPT}.
\newblock \url{https://openai.com/blog/chatgpt}, 2022.
\newblock Blog post published November 30, 2022.

\bibitem[\protect\citeauthoryear{Park \bgroup \em et al.\egroup }{2019}]{ParkJU:sofreggrfnldusre}
Jun-U Park, Sang-Ki Ko, Marco Cognetta, and Yo-Sub Han.
\newblock {{S}oft{R}egex: Generating Regex from Natural Language Descriptions using Softened Regex Equivalence}.
\newblock In {\em Proceedings of EMNLP-IJCNLP 2019}, pages 6425--6431, Hong Kong, China, November 2019.

\bibitem[\protect\citeauthoryear{Pavao \bgroup \em et al.\egroup }{2022}]{codalabcompetitions}
Adrien Pavao, Isabelle Guyon, Anne-Catherine Letournel, Xavier Baró, Hugo Escalante, Sergio Escalera, Tyler Thomas, and Zhen Xu.
\newblock {CodaLab Competitions: An open source platform to organize scientific challenges}.
\newblock {\em Technical report}, 2022.

\bibitem[\protect\citeauthoryear{Pitt and Warmuth}{1993}]{PittL:mincondpcbawap}
Leonard Pitt and Manfred~K. Warmuth.
\newblock {The Minimum Consistent DFA Problem Cannot Be Approximated within Any Polynomial}.
\newblock {\em J. ACM}, 40(1):95–142, jan 1993.

\bibitem[\protect\citeauthoryear{Radford \bgroup \em et al.\egroup }{2019}]{radford2019language}
Alec Radford, Jeffrey Wu, Rewon Child, David Luan, Dario Amodei, Ilya Sutskever, et~al.
\newblock Language models are unsupervised multitask learners.
\newblock {\em OpenAI blog}, 1(8):9, 2019.

\bibitem[\protect\citeauthoryear{Tenney \bgroup \em et al.\egroup }{2019}]{tenney2019bert}
Ian Tenney, Dipanjan Das, and Ellie Pavlick.
\newblock {BERT Rediscovers the Classical NLP Pipeline}.
\newblock In {\em Proceedings of the 57th Annual Meeting of the Association for Computational Linguistics}, pages 4593--4601, 2019.

\bibitem[\protect\citeauthoryear{Tunstall \bgroup \em et al.\egroup }{2023}]{Tunstall2023starchat-alpha}
Lewis Tunstall, Nathan Lambert, Nazneen Rajani, Edward Beeching, Teven Le~Scao, Leandro von Werra, Sheon Han, Philipp Schmid, and Alexander Rush.
\newblock Creating a coding assistant with starcoder.
\newblock {\em Hugging Face Blog}, 2023.
\newblock https://huggingface.co/blog/starchat.

\bibitem[\protect\citeauthoryear{Turing and others}{1936}]{turing1936computable}
Alan~Mathison Turing et~al.
\newblock On computable numbers, with an application to the entscheidungsproblem.
\newblock {\em J. of Math}, 58(345-363):5, 1936.

\bibitem[\protect\citeauthoryear{Vaandrager}{2017}]{10.1145/2967606}
Frits Vaandrager.
\newblock {Model Learning}.
\newblock {\em Commun. ACM}, 60(2):86–95, jan 2017.

\bibitem[\protect\citeauthoryear{Valizadeh and Berger}{2023}]{valizadeh2023search}
Mojtaba Valizadeh and Martin Berger.
\newblock {Search-Based Regular Expression Inference on a GPU}.
\newblock {\em Proc. ACM Program. Lang.}, 7(PLDI), jun 2023.
\newblock Draft available at \url{https://arxiv.org/abs/2305.18575}, implementation: \url{https://github.com/MojtabaValizadeh/paresy}.

\bibitem[\protect\citeauthoryear{van~der Poel \bgroup \em et al.\egroup }{2023}]{vanderpoel2023mlregtest}
Sam van~der Poel, Dakotah Lambert, Kalina Kostyszyn, Tiantian Gao, Rahul Verma, Derek Andersen, Joanne Chau, Emily Peterson, Cody~St. Clair, Paul Fodor, Chihiro Shibata, and Jeffrey Heinz.
\newblock {MLRegTest: A Benchmark for the Machine Learning of Regular Languages}, 2023.

\bibitem[\protect\citeauthoryear{Vaswani \bgroup \em et al.\egroup }{2017}]{vaswani2017attention}
Ashish Vaswani, Noam Shazeer, Niki Parmar, Jakob Uszkoreit, Llion Jones, Aidan~N Gomez, {\L}ukasz Kaiser, and Illia Polosukhin.
\newblock Attention is all you need.
\newblock {\em Advances in neural information processing systems}, 30, 2017.

\bibitem[\protect\citeauthoryear{Wang \bgroup \em et al.\egroup }{2019}]{wang2019superglue}
Alex Wang, Yada Pruksachatkun, Nikita Nangia, Amanpreet Singh, Julian Michael, Felix Hill, Omer Levy, and Samuel Bowman.
\newblock Superglue: A stickier benchmark for general-purpose language understanding systems.
\newblock {\em Advances in neural information processing systems}, 32, 2019.

\bibitem[\protect\citeauthoryear{Williams \bgroup \em et al.\egroup }{2018}]{williams2017broad}
Adina Williams, Nikita Nangia, and Samuel Bowman.
\newblock A broad-coverage challenge corpus for sentence understanding through inference.
\newblock In {\em Proceedings of the 2018 Conference of the North {A}merican Chapter of the Association for Computational Linguistics: Human Language Technologies, Volume 1 (Long Papers)}, pages 1112--1122, New Orleans, Louisiana, June 2018.

\bibitem[\protect\citeauthoryear{Yuie \bgroup \em et al.\egroup }{2021}]{wang2021codet5}
Wang Yuie, Weishi Wang, Shafiq Joty, and Steven~C.H. Hoi.
\newblock {CodeT5: Identifier-aware Unified Pre-trained Encoder-Decoder Models for Code Understanding and Generation}.
\newblock In {\em EMNLP}, 2021.

\bibitem[\protect\citeauthoryear{Zhong \bgroup \em et al.\egroup }{2018}]{ZhongZ:semregasbafgrefnls}
Zexuan Zhong, Jiaqi Guo, Wei Yang, Jian Peng, Tao Xie, Jian-Guang Lou, Ting Liu, and Dongmei Zhang.
\newblock {{S}em{R}egex: A Semantics-Based Approach for Generating Regular Expressions from Natural Language Specifications}.
\newblock In {\em Proceedings of EMNLP 2018}, pages 1608--1618, Brussels, Belgium, October-November 2018.

\end{thebibliography}

\end{document}


\maketitle

\section{Data Generation Hyperparameters}
We summarise the hyperparameters used for generating our datasets in Table~\ref{tab:data}. Symbols, operators (Ops) and abbreviations as specified in the main paper. ``Unif. CF'' is the uniform cost function, ``\#Rd. CF' is' the number of random cost functions per PN-set.
\begin{table}[h!]
    \small
	\center
	\begin{tabularx}{\columnwidth}{@{}l|YYYY@{}}
		\toprule
		& DS1 & DS2 & DS3 & DS4 \\
		\midrule
		$\Sigma$ & \multicolumn{2}{c}{$\epsilon+\{0,1\}$} & \multicolumn{2}{|c}{$\emptyset +\epsilon+\{0,1\} $}\\
		Ops & \multicolumn{2}{c}{$\cdot, +, ?, *$} & \multicolumn{2}{|c}{$\cdot, +, ?, *, \sim, -, \&$}\\
		\midrule
		\textsc{Type 1} \\
		p-range & \multicolumn{4}{c}{$1, \ldots, 10$}\\
		n-range & \multicolumn{4}{c}{$1, \ldots, 10$}\\
 		le-range & \multicolumn{4}{c}{$1, \ldots, 7$}\\
		\%Data & \multicolumn{4}{c}{${\sim}50$}\\
		\\
		\textsc{Type 2} \\
		p-range & \multicolumn{4}{c}{$1, \ldots, 10$}\\
		n-range & \multicolumn{4}{c}{$1, \ldots, 10$}\\
 		le-range & \multicolumn{4}{c}{$1, \ldots, 10$}\\
		\%Data & \multicolumn{4}{c}{${\sim}50$}\\
		\midrule
		Unif. CF & \checkmark & \checkmark & \checkmark & \checkmark\\
		\#Rd. CF & 0 & 19 & 0 & 19\\
		CF range & N/A & $1,\ldots,49$ & N/A & $1,\ldots,49$\\
		\bottomrule
	\end{tabularx}
    \caption{Data generation hyperparameters.}
    \label{tab:data}
\end{table}

\section{ReGPT Hyperparameters}
For ReGPT we train two models from scratch, on the training splits of DS~1+DS~2, and DS~3+DS~4 respectively.
We use the same hyperparameters for each, for training and inference, shown in Table~\ref{tab:regpt}.
\begin{table}[h!]
    \small
	\center
	\begin{tabularx}{.75\columnwidth}{@{}lc@{}}
	\toprule
	\multicolumn{2}{c}{Model}\\
	Vocab Size & 132 \\
	Context Length & 256 \\
	Embedding dims & 1,024 \\
	Layers & 24 \\
	Attention Heads & 16\\
	\#Parameters & ${\sim}$300M\\
	\midrule
	\multicolumn{2}{c}{Training}\\
    Training GPUs & 4xV100 16GB\\
	Training Time & ${\sim}12$ hours/model\\
	Optimiser & AdamW\\
	Batch Size & 512\\
	LR & $1e^{-5}$\\
	Weight Decay & $0.1$ \\
	Warmup Steps & 500 \\
	LR Scheduler & Cosine \\
	Epochs & 100\\
	\midrule
	\multicolumn{2}{c}{Inference}\\
    Inference GPUs & 4xV100 16GB\\
	Sampling method & Nucleus Sampling\\
	top-p & 0.8\\
	temperature & 0.8\\
	\#Samples & $10$ or $1,000$ (incl. $1$ greedy)\\
	\bottomrule
	\end{tabularx}
 \caption{ReGPT hyperparameters.}
 \label{tab:regpt}
\end{table}

\section{StarChat$\pmb{\beta}$ Hyperparameters}
In Table~\ref{tab:starchat} we give the hyperparameters used for inference with the StarChat$\beta$ few-shot prompting baseline.

Prompts used for Datasets 1 and 2, and Datasets 3 and 4 respectively are shown in their own subsections below; expressions in curly brackets $\{\ldots\}$ are instantiated with postive/negative sets and cost function values per problem.

We stop generation after the first linebreak character (\textbackslash n) is produced, using tokens after ``\texttt{r = }'' as the expression.
\begin{table}[h!]
    \small
	\center
	\begin{tabularx}{.79\columnwidth}{@{}lc@{}}
	\toprule
	\multicolumn{2}{c}{Model}\\
	Vocab Size & 49,156 \\
	Context Length & 8,192 \\
	Embedding dims & 6,144 \\
    Inner dims & 24,576 \\
	Layers & 40 \\
	Attention Heads & 48\\
	\#Parameters & ${\sim}$16B\\
	\midrule
	\multicolumn{2}{c}{Inference}\\
    Inference GPUs & 4xV100 16GB\\
	Sampling method & Prompting + Nucleus Sampling\\
	top-p & 0.8\\
	temperature & 0.8\\
	\#Samples & 10\\
	\bottomrule
	\end{tabularx}
 \caption{StarChat$\beta$ hyperparameters.}
 \label{tab:starchat}
\end{table}

\pagebreak
\section{ReGPT Encoding}
\begin{figure*}[t!]
\center
{
{\small
\begin{tabularx}{\textwidth}{lX}
Positive Set: & $11, 0000, 000$\\
Negative Set: & $\epsilon, 1, 101$\\
RE Characters \& Ops: &  $\{\emptyset, \epsilon, a, ?, *, \sim, \cdot, \&, +, -\} $\\
Cost function: & \text{uniform}\\
Input Encoding: & [CLS] [POS] 1 1 [POS] 0 0 0 0 [POS] 0 0 0 [NEG] e [NEG] 1 [NEG] 1 0 1 [COST\_A] 1 [COST\_?] 1 [COST\_*] 1 [COST\_.] 1 [COST\_+] 1 [BOR]  ( 0 * ) . ( 0 + ( 1 . 1 ) ) [EOR]
\end{tabularx}}
}
\caption{Example of ReGPT encoding.}
\label{fig:encoding}
\end{figure*}
Figure~\ref{fig:encoding} shows and example of the input encoding for our ReGPT model, for a training instance from Dataset 1. Note that we use COST\_A to cover all characters $a\in\Sigma$, $\epsilon$, and $\emptyset$. To stay with ASCII characters, we encode $\epsilon$ as 'e', $\emptyset$ as 'E', and concatenation ($\cdot$) as '.' In practice, we also distinguish between $0$ and $1$ characters used for PN-strings and regular expressions from integers used for the cost function, by encoding the former as tokens 'ONE' and 'ZERO'.

\onecolumn
\subsection{StarChatt$\pmb{\beta}$ Prompt, Datasets 1 and 2}
{\small
\texttt{
\\
<|system|>\\
<|end|>\\
<|user|>\\
Regular Expression Inference is the task of finding a minimal regular expression r with respect to a cost function c, that accepts all strings in a positive set p and rejects all strings in a negative set n.\\
\\
The alphabet for regular expressions is {0, 1}, and the empty string e. The empty set is represented as E.\\
\\
Operators for regular expressions are Option (?), Kleene-Star (*), Concatenation (.), and Union (+).
Parentheses are allowed, and are not used when calculating cost.\\
\\
The cost of a regular expression r is the sum of costs of all symbols and operators used in r, according to the cost function c.\\
\\
Examples:\\
Input:\\
p = [0101]\\
n = [1000100]\\
c(0) = 1\\
c(1) = 1\\
c(e) = 1\\
c(E) = 1\\
c(?) = 1\\
c(*) = 1\\
c(.) = 1\\
c(+) = 1\\
Output:\\
r = ((0.1)*)\\
cost = 4\\
\\
Input:\\
p = [10, 000, 1101110, 1000000, 1110110, 010]\\
n = [000110, 0110, 01101000]\\
c(0) = 1\\
c(1) = 1\\
c(e) = 1\\
c(E) = 1\\
c(?) = 1\\
c(*) = 1\\
c(.) = 1\\
c(+) = 1\\
Output:\\
r = (((1.(0*))*).((0.(1?))*))\\
cost = 11\\
\\
Input:\\
p = [010011]\\
n = [000000 00011 110010 111010]\\
c(0) = 20\\
c(1) = 20\\
c(e) = 20\\
c(E) = 20\\
c(?) = 8\\
c(*) = 3\\
c(.) = 45\\
c(+) = 38\\
Output:\\
r = (0.((1.(0*))*))\\
cost = 156\\
\\
Find the minimal regular expression r for the inputs below. Only answer with the final regular expression:\\
Input:\\
p = \{p\}\\
n = \{n\}\\
c(0) = \{c0\}\\
c(1) = \{c1\}\\
c(e) = \{ce\}\\
c(E) = \{cE\}\\
c(?) = \{cq\}\\
c(*) = \{cs\}\\
c(.) = \{cd\}\\
c(+) = \{cp\}\\
Output:\\
<|end|>\\
<|assistant|>\\
r = 
}
}

\subsection{StarChat$\pmb{\beta}$ Prompt, Datasets 3 and 4}
{\small
\texttt{
\\
<|system|>\\
<|end|>\\
<|user|>\\
Regular Expression Inference is the task of finding a minimal regular expression r with respect to a cost function c, that accepts all strings in a positive set p and rejects all strings in a negative set n.\\
\\
The alphabet for regular expressions is {0, 1}, and the empty string e. The empty set is represented as E.\\
\\
Operators for regular expressions are Option (?), Kleene-Star (*), Concatenation (.), Union (+), Complement ($\sim$), Intersection (\&), and Restriction (-).\\
\\
Parentheses are allowed, and are not used when calculating cost.\\
\\
The cost of a regular expression r is the sum of costs of all symbols and operators used in r, according to the cost function c.\\
\\
Examples:\\
Input:\\
p = [0101]\\
n = [1000100]\\
c(0) = 1\\
c(1) = 1\\
c(e) = 1\\
c(E) = 1\\
c(?) = 1\\
c(*) = 1\\
c(.) = 1\\
c(+) = 1\\
c($\sim$) = 1\\
c(\&) = 1\\
c(-) = 1\\
Output:\\
r = ((0.1)*)\\
cost = 4\\
\\
p = [0, 11, 011, 110, 10]\\
n = [e, 00, 000, 010, 1, 100, 101]\\
c(0) = 1\\
c(1) = 1\\
c(e) = 1\\
c(E) = 1\\
c(?) = 1\\
c(*) = 1\\
c(.) = 1\\
c(+) = 1\\
c($\sim$) = 1\\
c(\&) = 1\\
c(-) = 1\\
r = (~((1?).(((0.(1?))*)-0)))\\
cost = 11\\
\\
Input:\\
p = [011, 0, 1, 101]\\
n = [e, 10, 100, 11, 110]\\
c(0) = 1\\
c(1) = 1\\
c(e) = 1\\
c(E) = 1\\
c(?) = 36\\
c(*) = 20\\
c(.) = 38\\
c(+) = 1\\
c($\sim$) = 10\\
c(\&) = 12\\
c(-) = 30\\
r = (0+((~1).1))\\
52\\
\\
\\
Find the minimal regular expression r for the inputs below. Only answer with the final regular expression:\\
Input:\\
p = \{p\}\\
n = \{n\}\\
c(0) = \{c0\}\\
c(1) = \{c1\}\\
c(e) = \{ce\}\\
c(E) = \{cE\}\\
c(?) = \{cq\}\\
c(*) = \{cs\}\\
c(.) = \{cd\}\\
c(+) = \{cp\}\\
c($\sim$) = \{ct\}\\
c(\&) = \{ca\}\\
c(-) = \{cm\}\\
Output:\\
<|end|>\\
<|assistant|>\\
r = 
}